\documentclass{article}

\usepackage[preprint]{neurips_2026}  
\makeatletter
\renewcommand{\@noticestring}{arXiv preprint.}
\makeatother

\usepackage[utf8]{inputenc}
\usepackage[T1]{fontenc}
\usepackage{url}
\usepackage{booktabs}
\usepackage{amsfonts}
\usepackage{amsmath}
\usepackage{amssymb}
\usepackage{nicefrac}
\usepackage{microtype}
\usepackage{graphicx}
\usepackage{subcaption}
\usepackage{algorithm}
\usepackage{algorithmic}
\usepackage{multirow}
\usepackage{makecell}
\usepackage{wrapfig}

\usepackage[dvipsnames,svgnames]{xcolor}
\usepackage{hyperref}
\hypersetup{
  colorlinks=true,
  linkcolor=magenta,
  citecolor=ForestGreen,
  urlcolor=blue!80!black,
  filecolor=blue!80!black,
  }
  
\usepackage[capitalize,noabbrev]{cleveref}

\newcommand\blfootnote[1]{%
  \begingroup
  \renewcommand\thefootnote{}\footnote{#1}%
  \addtocounter{footnote}{-1}%
  \endgroup
}

\title{Rethinking Positional Encoding for\\ Neural Vehicle Routing}

\author{%
  Chuanbo Hua$^{1}${\mdseries,}\ \ 
  Federico Berto$^{2}${\mdseries,}\ \ 
  André Hottung$^{3}${\mdseries,}\ \ 
  Nayeli Gast Zepeda$^{4}${\mdseries,} \\
  \bf Yining Ma$^{5}${\mdseries,}\ \ 
  Zihan Ma$^{1}${\mdseries,}\ \ 
  Paula Wong-Chung$^{6}${\mdseries,}\ \ 
  Changhyun Kwon$^{1}${\mdseries,} \\
  \bf Cathy Wu$^{5}${\mdseries,}\ \ 
  Kevin Tierney$^{4}${\mdseries,}\ \ 
  Jinkyoo Park$^{1}$ \\[0.5em]
  \normalfont\small
  $^{1}$KAIST \quad
  $^{2}$Radical Numerics \quad
  $^{3}$Bielefeld University \quad
  $^{4}$University of Vienna \\
  $^{5}$MIT \quad
  $^{6}$University of British Columbia \quad
  AI4CO\thanks{Works from the AI4CO open research community.}
}

\renewcommand{\thefootnote}{\fnsymbol{footnote}}
\begin{document}
\maketitle

\blfootnote{The implementation is open-sourced at \url{https://github.com/ai4co/rl4co}.}

\renewcommand{\thefootnote}{\arabic{footnote}}
\setcounter{footnote}{0}

\vspace{-6mm}

\begin{abstract}
Transformer-based models have become the dominant paradigm for neural combinatorial optimization (NCO) of vehicle routing problems (VRPs), yet the role of positional encoding (PE) in these architectures remains largely unexplored.
Unlike natural language, where tokens are uniformly spaced on a line, routing solutions exhibit several properties that render standard NLP positional encodings inadequate. In this work, we formalize three such structural properties that a routing-aware PE should respect, namely anisometric node distances, cyclic and direction-aware topology, and hierarchical depot-anchored global multi-route structure, combining them with a unifying design principle of geometric grounding.
Guided by these criteria, we analyze and compare PE methods spanning NLP, graph-transformer, and routing-specific families, and propose a hierarchical anisometric PE that combines a distance-indexed, circularly consistent in-route encoding with a depot-anchored angular cross-route encoding.
Extensive experiments across diverse VRP variants demonstrate that geometry-grounded PE consistently outperforms index-based alternatives, with gains that transfer across problem variants, model architectures, and distribution shifts.
\end{abstract}
\section{Introduction}
\label{sec:introduction}

Neural combinatorial optimization (NCO) has emerged as a promising paradigm for vehicle routing problems (VRPs), with learning-based solvers steadily narrowing the gap to classical operations-research methods~\citep{kool2018attention,kwon2020pomo,ma2021learning,drakulic2024bq,luo2023neural,hottung2025neural}.
A defining challenge in this setting is that routing inputs are unordered sets of nodes, while routing outputs are highly structured sequences, including tours, multi-route solutions, and ordering chains.
Bridging this gap requires the model to reason explicitly about \emph{position}: which node comes next, how far it is from its neighbors in the current solution, and which route it belongs to.
Transformer-based architectures, equipped with \emph{positional encoding} (PE) as the standard mechanism for injecting such positional information, have become the dominant backbone, and a series of methods~\citep{kool2018attention,kwon2020pomo,ma2021learning,ma2022efficient,drakulic2024bq,luo2023neural,hottung2025neural} have progressively closed the quality gap with classical solvers.

Despite this rapid progress, the role of PE has received remarkably little scrutiny.
Most construction-based models omit PE entirely~\citep{kool2018attention,kwon2020pomo} or directly apply standard natural language processing (NLP) encodings without adaptation.
Improvement-based methods have begun to incorporate positional signals, such as a Gray-code cyclic PE in DACT~\citep{ma2021learning} and an index-based PE in NDS~\citep{hottung2025neural}, but these designs are ad hoc, neither justified by the structure of routing solutions nor compared against alternatives.
In contrast, PE design in NLP and vision has received sustained attention and has been shown to meaningfully affect model performance~\citep{vaswani2017attention,shaw2018self,su2024roformer}.

This neglect matters because routing solutions live in a positional space that is richer than the linear, isometric sequence assumed by NLP PE.
While NLP positions are uniformly spaced on a line, routing positions exhibit structure at three distinct levels.
At the metric level, routes are \emph{anisometric}: consecutive nodes are separated by varying physical distance rather than equal index gaps, so a sinusoidal encoding that maps index pairs to identical embedding gaps ignores the geometry of the underlying tour.
At the topological level, routes are \emph{cyclic}, since each closed tour departs from and returns to the depot, making the first and last positions topologically identical, a property no linear PE can express.
Routes are also \emph{direction-aware} in a problem-dependent way: symmetric VRPs admit reversal symmetry, whereas PDTSP and VRPTW carry precedence and time-window semantics that break it.
At the global level, VRP solutions are \emph{hierarchical} and \emph{depot-anchored}: a node's identity depends on its intra-route position, on which route it belongs to, and on how that route relates to others through a structurally distinguished depot.
Moreover, the set of routes admits no canonical ordering, so a hierarchical PE must be permutation-invariant across routes while sequential within them, a structure that no single one-dimensional positional index can capture.

Building on these observations, this paper presents the first systematic study of positional encoding for neural vehicle routing.
Our contributions are as follows.
\begin{itemize}
\item We identify the structural properties a positional encoding should respect for vehicle routing, and use them to organize existing encodings into a unified taxonomy.
\item We propose a hierarchical positional encoding tailored to the geometry of routing solutions.
\item We show through probing and controlled comparison that existing encodings miss these properties and that this gap propagates to solution quality.
\item We demonstrate consistent gains across problem variants, architectures, and distribution shifts when plugged into state-of-the-art solvers.
\end{itemize}

\begin{figure}[t]
\centering
\includegraphics[width=\textwidth]{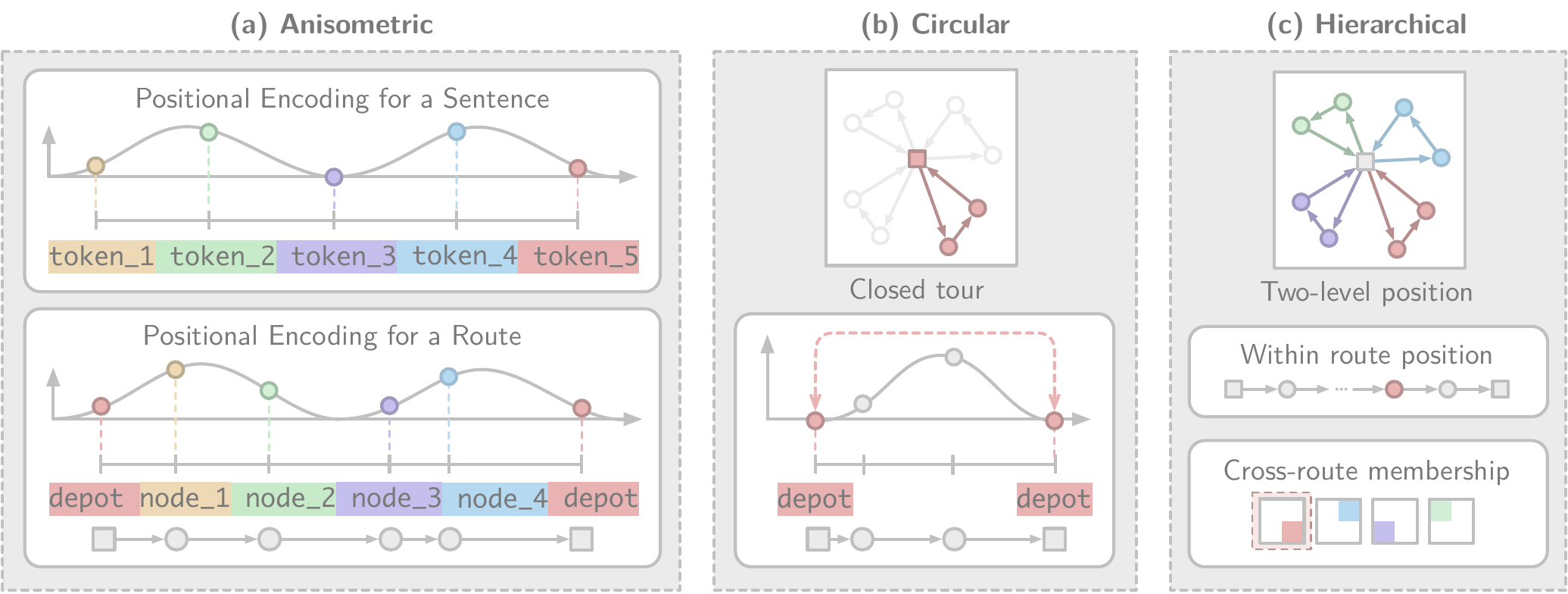}
\caption{Properties of positional encoding for routes.}
\label{fig:teaser}
\vspace{-5mm}
\end{figure}
\section{Background and Related Work}
\label{sec:background}

\paragraph{Vehicle routing problems.}
A VRP is defined on a graph $G = (V, E)$, where $V = \{0, 1, \dots, N\}$ contains a depot $x_0$ and $N$ customers, and each edge carries a travel cost (typically Euclidean distance).
A feasible solution partitions customers into routes, each originating from and returning to the depot, subject to variant-specific constraints such as vehicle capacity (CVRP), time windows (VRPTW), or optional service with prizes (PCVRP).
The objective is to minimize total travel cost (or cost minus collected prizes in PCVRP).
We defer full mathematical formulations to \cref{sec:appendix-vrp-def}.

\paragraph{Transformer-based NCO.}
Neural VRP solvers fall into two families with fundamentally different relationships to PE.
\emph{Construction methods} \citep{kool2018attention,kwon2020pomo,drakulic2024bq,luo2023neural} decode one node at a time from a permutation-equivariant encoder over an unordered customer set, so PE is not needed at the encoding stage; re-encoding variants such as LEHD and BQ are an exception, since the partial solution they re-encode is an ordered sequence whose positional structure current designs are unexplored.
\emph{Improvement methods} \citep{ma2021learning,ma2022efficient,hottung2025neural} operate on a complete current solution and refine it through learned $k$-opt moves, neighborhood search, or large-neighborhood deconstruction, so every forward pass consumes an ordered sequence of routed nodes and PE is an explicit, unavoidable design choice.
Our analysis therefore concentrates on improvement methods and on re-encoding-style construction methods, where PE materially shapes the representation.

\paragraph{Positional encoding in routing models.}
The treatment of PE in existing routing models is fragmented and largely unexamined.
Most construction models in the encoder do not use PE at all \citep{kool2018attention,kwon2020pomo}, relying instead on the permutation-equivariant nature of self-attention over node coordinates.
Among improvement models, DACT \citep{ma2021learning} introduces a cyclic PE (CPE) based on Gray codes to represent the circular ordering of nodes within a tour.
NDS \citep{hottung2025neural} uses a simple index-based PE to encode the sequential position of each node in its route.
CycleFormer \citep{yook2024cycleformer} proposes a circular PE specifically designed for TSP.
However, none of these works systematically compare their PE choice against alternatives, nor do they articulate what properties a PE should have for routing.
This paper fills that gap.

\section{Routing-Aware Positional Encoding Taxonomy}
\label{sec:taxonomy}

This section examines the structural properties of routing solutions and existing PE designs.
\Cref{subsec:preliminaries} formalizes the geometric properties of routes that we will reason about and states desiderata that follow from them.
\Cref{subsec:taxonomy} surveys existing PE methods, and discusses at the conceptual level the extent to which each satisfies the desiderata.

For clarity of exposition, the analysis throughout this section assumes \emph{symmetric} distances and \emph{closed-tour} VRP variants, i.e., every route vehicle returns to the depot.
The treatment of asymmetric and open-tour variants is presented in \cref{sec:discussion}.

\subsection{Preliminaries and Desiderata}
\label{subsec:preliminaries}

A solution of a VRP variant is a collection of routes $\mathcal{S} = \{r^{(1)}, \dots, r^{(K)}\}$, where each route $r^{(k)} = (v_1^{(k)}, \dots, v_{L_k}^{(k)})$ visits a subset of customers, with $v_1^{(k)} = v_{L_k}^{(k)} = x_0$, and $x_0$ being the depot.
We highlight three structural properties of such solutions and motivate the corresponding desiderata.

\begin{wraptable}{r}{0.45\textwidth}
\vspace{-1.0\baselineskip}
\centering
\small
\caption{Anisometry measurement of near-optimal routes across VRP variants.}
\label{tab:anisometry}
\begin{tabular}{lccc}
\toprule
Problem & CV & max/min &  MAD \\
\midrule
CVRP-100  & 0.81 & 26.6 & 0.37  \\
CVRP-200  & 0.93 & 42.1 & 0.44  \\
CVRP-500  & 1.12 & 72.3 & 0.53  \\
VRPTW-100 & 0.78 & 24.0 & 0.33  \\
PDTSP-51    & 0.78 & 68.9 & 1.24 \\
\bottomrule
\end{tabular}
\vspace{-1.0\baselineskip}
\end{wraptable}

\paragraph{Anisometry.}
The Euclidean distances between consecutive nodes $\|x_{v_{i+1}} - x_{v_i}\|_2$ generally vary with $i$ within a route, so the cumulative distance $d_i = \sum_{j=2}^{i} \|x_{v_j} - x_{v_{j-1}}\|_2$ is a nonlinear function of the discrete index.
Equal increments in index do not correspond to equal increments in physical distance.
Standard positional encodings that map uniform indices to embedding space implicitly assume isometric spacing and discard this geometric information.
\Cref{tab:anisometry} confirms that this anisometry is pronounced in near-optimal solutions across VRP variants.
For each route in $1{,}000$ instances per problem solved with LKH/HGS \citep{helsgaun2017extension,vidal2022hybrid}, we report the coefficient of variation (CV) of consecutive edge lengths, the max-to-min edge-length ratio (max/min), and the residual median absolute deviation (MAD) of a linear fit of $d_i$ against $i$ in units of the average edge length.
The CV is large in every setting and grows monotonically with problem size on CVRP, indicating that edge-length heterogeneity scales with the instance rather than vanishing as a finite-sample artifact.
Edge-length ratios reach the tens to hundreds, and the residual MAD shows that a linear index-to-distance fit deviates by roughly a third to a half of an average edge length on the multi-route variants.\footnote{The PDTSP-51 row reports a larger residual MAD because the problem is single-tour: the linear fit is driven by a small number of long routes, which inflates the MAD relative to the average edge.}

\paragraph{Circularity.}
Under the closed-tour assumption, the first and last positions of every route are the same physical node yet any PE built from a linear index assigns these two positions the maximum possible embedding distance.
Two further observations could refine this property.
For variants with directional constraints such as VRPTW, the cyclic ordering is itself directional: traversing a route in reverse changes feasibility, so the encoding must preserve direction in order for solution validity to be retained.
For variants without such constraints, the same route traversed in either direction has the same cost, so a directionally-invariant encoding generalizes more readily across equivalent solution representations.
A routing-aware PE should respect both regimes.

\vspace{-2mm}

\paragraph{Hierarchy.}
A multi-route solution is a two-level object: a node's identity is determined jointly by its within-route position and its route membership.
A one-dimensional positional index conflates these two levels, so the ``second node of route A'' and the ``second node of route B'' receive identical encodings even though route membership itself carries information about the node's role in the global solution.
We quantify the resulting ambiguity by collecting the set of customers occupying position $k$ across the routes of an instance and measuring the entropy of their depot-relative polar-angle distribution.
Across CVRP-\{100, 200, 500\} and VRPTW-100, the normalized entropy averaged over $k = 2, \dots, 15$ is $0.992$, with a minimum of $0.953$ over the entire (problem, $k$) grid: the within-route index $k$ retains at most $4.7\%$ of the angular information about a node's position relative to the depot.
A PE that maps the abstract index $k$ into a node embedding therefore cannot well distinguish nodes that share $k$ but lie in different parts of the instance.

Based on previous properties, we identify three structural properties of routing solutions that a positional encoding should respect  as desiderata.

\textit{Distance-awareness} (D1):
the PE of a node should reflect its cumulative travel distance along the route rather than its discrete index;
formally, for nodes $v_i, v_j$ on the same route, $\|\mathrm{PE}(v_i) - \mathrm{PE}(v_j)\|$ should correlate with $|d_i - d_j|$ rather than $|i - j|$.

\textit{Circularity} (D2):
for closed-tour variants, the PE should satisfy $\mathrm{PE}(v_1) = \mathrm{PE}(v_L)$, and its directional/non-directional behavior should match the variant's symmetry.

\textit{Hierarchical structure} (D3):
the PE should encode both within-route position and across-route membership, disambiguating nodes that share a within-route index but belong to different routes.


\subsection{A Taxonomy of Existing PE Methods}
\label{subsec:taxonomy}

We organize existing PE methods into three categories by their design origin and discuss, for each method, which of desiderata are satisfied at the conceptual level.

\vspace{-2mm}

\begin{wraptable}{r}{0.45\textwidth}
\vspace{-1.0\baselineskip}
\centering
\small
\caption{Conceptual satisfaction of the routing PE desiderata by existing methods. \checkmark: fully satisfied; $\sim$: partially satisfied; $\times$: not satisfied.}
\label{tab:desiderata}
\begin{tabular}{lccc}
\toprule
Method & D1 & D2 & D3 \\
\midrule
APE              & $\times$ & $\times$   & $\times$ \\
Sinusoidal (SIN) & $\times$ & $\times$   & $\times$ \\
RoPE             & $\times$ & $\sim$     & $\times$ \\
RPE              & $\times$ & $\sim$     & $\times$ \\
ALiBi            & $\times$ & $\times$   & $\times$ \\
Laplacian PE     & $\times$ & $\times$   & $\sim$   \\
RWSE             & $\times$ & $\times$   & $\sim$   \\
SPD              & $\sim$   & $\times$   & $\times$ \\
DACT CPE         & $\times$ & \checkmark & $\times$ \\
CycleFormer PE   & $\times$ & \checkmark & $\times$ \\
\bottomrule
\end{tabular}
\vspace{-1.0\baselineskip}
\end{wraptable}

\paragraph{NLP-Originated PE.}
These encodings were designed for sequential text data and assume uniform token spacing: 
\textit{Absolute PE (APE)} is a learnable embeddings indexed by position in the sequence \citep{gehring2017convolutional}.
Each position $i$ maps to a trainable vector $\mathbf{e}_i \in \mathbb{R}^D$.
APE does not satisfy any of D1-D3: it assumes uniform spacing, is not circular, and has no hierarchical component.
\textit{Sinusoidal PE (SIN)} fixed sinusoidal functions of the integer index \citep{vaswani2017attention}.
Like APE, SIN maps uniform indices and does not satisfy D1-D3.
\textit{Rotary PE (RoPE)} encodes positions via rotation matrices applied to query-key pairs \citep{su2024roformer}.
The attention score between positions $i$ and $j$ depends only on $i - j$, providing relative position sensitivity.
RoPE partially satisfies D2 through its multiplicative structure but does not satisfy D1 or D3.
\textit{Relative PE (RPE)} adds learnable bias terms to attention logits based on the relative offset $i - j$ \citep{shaw2018self}.
Similar to RoPE, RPE captures relative ordering but assumes uniform spacing and lacks both full circularity and hierarchy.
\textit{ALiBi} subtracts a linear penalty $m \cdot |i - j|$ from attention logits, where $m$ is a head-specific slope \citep{press2021train}.
ALiBi is simple and extrapolation-friendly but does not satisfy D1--D3.

The standard NLP encodings are likely to fail to satisfy the desiderata because such encodings are parameterized by \emph{abstract index sequences}, which carry no information about the underlying geometry of the solution.

\paragraph{Graph Transformer PE.} These encodings exploit graph-structural information and are used in graph transformers:
\textit{Laplacian PE (Lap.\ PE)} uses the $k$ smallest non-trivial eigenvectors of the graph Laplacian as node features \citep{dwivedi2020generalization}.
Laplacian PE captures global graph structure and partially satisfies D3 (nodes in the same cluster receive similar embeddings) but does not encode route-level distance (D1) or circularity (D2).
\textit{Random Walk SE (RWSE)} encodes each node by the diagonal of the random walk matrix powers $\{RW^k\}_{k=1}^{K}$, capturing local neighborhood structure \citep{dwivedi2021graph}.
RWSE partially satisfies D3 through local structure encoding but lacks D1 and D2.
\textit{Shortest-Path Distance (SPD)} uses the shortest-path distance between node pairs as a bias in attention \citep{ying2021do}.
SPD partially satisfies D1 (distance-awareness through graph distances) but does not address D2 or D3 in the routing context.

\paragraph{Routing-Specific PE.} These encodings are designed with routing structure in mind.
\textit{DACT CPE} is the cyclic PE introduced by DACT \citep{ma2021learning} and uses a cyclic input sequence representation to assign position encoding, satisfying D2 (circularity).
However, it does not encode actual travel distances (D1) and lacks hierarchical cross-route information (D3).
\textit{CycleFormer Circular PE} is designed specifically for TSP and constructs a circular PE that wraps position indices around the tour \citep{yook2024cycleformer}.
It satisfies D2, but, like DACT CPE, does not address D1 or D3.

In summary, \Cref{tab:desiderata} shows that no existing PE method satisfies all three desiderata: NLP-originated encodings address none, or at most partially address D2, graph transformer encodings partially address D3 but ignore the routing-specific properties, and existing routing-specific PEs achieve circularity but neither distance-awareness nor hierarchy.

\section{Hierarchical Positional Encoding}
\label{sec:method}

This section introduces a positional encoding constructed to satisfy
the desiderata D1--D3 of \cref{subsec:preliminaries}.
The encoding consists of two complementary components defined per node:
an \emph{in-route positional encoding} (IPE) capturing the
distance-indexed, circular ordering of a node within its route, and a
\emph{cross-route positional encoding} (XPE) capturing the
depot-anchored angular position of the node within the global solution.


\subsection{In-Route Component: IPE}
\label{subsec:ipe}

\paragraph{Distance-indexed sinusoidal encoding.}
For a route $r = (v_1, \dots, v_L)$ and
$v_1 = v_L = x_0$, define the cumulative travel distance along the
route,
\begin{equation*}
d_i = \sum_{j=2}^{i} \big\| x_{v_j} - x_{v_{j-1}} \big\|_2,
\qquad d_L \;=\; \text{total route length} > 0.
\end{equation*}
We rescale this quantity to a single sinusoidal period,
$\hat{d}_i \;=\; 2\pi \, d_i / d_L \;\in\; [0, 2\pi]$,
and compose a multi-frequency encoding using a geometric sequence of
frequencies $\omega_k = \lambda^{-2k/D}$ for $k = 0, \dots, D/2 - 1$,
with $\lambda = 10{,}000$ following~\citet{vaswani2017attention}.
The construction admits two variants, selected to match the symmetry
of the VRP variant being solved.

\paragraph{Direction-aware variant.}
For variants whose feasibility is sensitive to traversal direction
(VRPTW, PDTSP, and any problem with precedence or time-window constraints),
we use the standard sine--cosine pair,
\begin{equation*}
\mathrm{IPE}^{\rightarrow}(v_i) \;=\; \big[\sin(\omega_k \hat{d}_i),\, \cos(\omega_k \hat{d}_i)\big]_{k=0}^{D/2-1}.
\end{equation*}
Reversing the route maps $d_i \mapsto d_L - d_i$, which flips the sign
of every sine component and leaves the cosine components unchanged, so
the encoding distinguishes a route from its reversal.

\paragraph{Direction-invariant variant.}
For reversal-symmetric variants (CVRP, TSP), we retain only the cosine
components,
\begin{equation*}
\mathrm{IPE}^{\leftrightarrow}(v_i) \;=\; \big[\cos(\omega_k \hat{d}_i)\big]_{k=0}^{D/2-1},
\end{equation*}
doubling the number of cosine frequencies to keep the embedding
dimension fixed.
Because $\cos(2\pi - x) = \cos(x)$, this encoding is invariant to route
reversal by construction, and the directional clause of D2 is satisfied
without any architectural symmetrization (in contrast to, e.g., DACT, which augments training with reversed tours).

\paragraph{Satisfaction of D1 and D2.}
The encoding is a function of cumulative travel distance rather than
discrete index, so the embedding gap between two nodes on the same
route is governed by their physical separation along the tour,
satisfying D1.
The rescaling maps both endpoints of a
closed tour to the same phase, so $\mathrm{IPE}(v_1) = \mathrm{IPE}(v_L)$
for both variants, satisfying the topological clause of D2.

\subsection{Cross-Route Component: XPE}
\label{subsec:xpe}

\paragraph{Per-node depot-anchored polar encoding.}
For each customer $v$ with coordinate $x_v \in \mathbb{R}^2$, define the
depot-anchored polar angle
\begin{equation*}
\theta_v \;=\; \mathrm{atan2}~\!\big(x_v^{(y)} - x_0^{(y)},\; x_v^{(x)} - x_0^{(x)}\big) \;\in\; [-\pi, \pi),
\end{equation*}
and encode it via a multi-frequency sinusoidal map of the same form as
IPE,
\begin{equation*}
\mathrm{XPE}(v) \;=\; \big[\sin(\omega_k' \theta_v),\, \cos(\omega_k' \theta_v)\big]_{k=0}^{K-1},
\qquad \omega_k' = 2^{k},
\end{equation*}
where the number of frequency bands $K$ is chosen so that
$2K \leq D$; the remaining dimensions of XPE are zero-padded to match
the IPE width.

\paragraph{Choice of polar over Cartesian.}
We encode $\theta_v$ rather than the Cartesian offset
$(x_v - x_0,\, y_v - y_0)$ because the polar angle is bounded and
periodic, matching the sinusoidal encoding form used by IPE; and
because the angular-sector structure of near-optimal VRP solutions
(\cref{sec:taxonomy}) makes the angular coordinate the geometrically
meaningful one, where routes are organized around the depot by direction,
not by signed displacement.

\paragraph{Why per node rather than per route.}
A natural alternative would be to summarize each route into a single
angular phase, e.g. the arithmetic, circular, or demand-weighted mean of
$\{\theta_v\}_{v \in r}$, or its median, and then broadcast this scalar to
all nodes on the route.
We avoid this design because no such aggregation is canonical,
each candidate is ad hoc, and the per-node encoding strictly dominates
any aggregate in information content: the multiset
$\{\theta_v\}_{v \in r}$ recovers any aggregation statistic, but not
conversely.
The per-node design therefore preserves full angular content and leaves
any pooling to be learned downstream by attention.

\paragraph{Satisfaction of D3.}
Two nodes at the same within-route position in different routes
receive nearly identical IPE but differ in XPE whenever
$\theta_v \neq \theta_w$, so the joint encoding separates them.
This is empirically supported by \cref{subsec:preliminaries}: the
depot-anchored polar-angle distribution conditioned on within-route
index has normalized entropy $0.992$, indicating that $\theta_v$ and
the within-route index are nearly independent and thus contribute
largely orthogonal signals.

\paragraph{Robustness to angular overlap.}
The disambiguation argument relies on routes occupying largely
non-overlapping angular sectors, which holds for near-optimal solutions
but degrades under heavy capacity constraints, on out-of-distribution
instances, or during early training when the current solution is far
from optimal.
When two routes overlap angularly, XPE alone no longer separates
same-index nodes; in such cases IPE provides the residual disambiguation
through the within-route distance signal, and the joint encoding remains
informative as long as the two nodes do not additionally coincide in
fractional within-route position.
\Cref{sec:experiments} reports robustness experiments under varying
capacity ratios and distribution shifts that probe this regime
empirically.

\subsection{Integration into the Encoder}
\label{subsec:integration}

The two positional components are concatenated per node and projected
jointly with the raw coordinate by a shared feed-forward layer,
\begin{equation*}
h_v^{(0)} \;=\; \mathrm{FF}~\!\big([\,x_v\ \|\ \mathrm{IPE}(v)\ \|\ \mathrm{XPE}(v)\,]\big),
\end{equation*}
and the resulting embeddings are passed to a standard Transformer
encoder without further modification.
We adopt concatenation rather than additive PE because the raw
coordinate and the sinusoidal positional signals live on different
geometric and numerical scales, and a joint MLP projection learns the
appropriate fusion.
We deliberately avoid any specialized cross-route mechanism in the
architecture so that the hierarchical structure of the encoding lives
entirely in the PE itself, ensuring that comparisons against PE
baselines (\cref{sec:experiments}) isolate the effect of the encoding
rather than entangling it with architectural changes.

\section{Experiments}
\label{sec:experiments}

The experiments are organized along three layers.
Layer 1 addresses the question \emph{``how much does the PE itself matter''}. It isolates the encoding under a fixed lightweight backbone at small instance sizes, where running ten-plus encodings to convergence is feasible.
Layer 2 addresses the question \emph{``can our PE push the strongest current method further''}. It plugs the encoding into NDS \citep{hottung2025neural} to compare against strongest operations-research and neural baselines.
Layer 3 addresses the question \emph{``does the encoding transfer across architectures and distributions''}. It plugs the encoding into N2S \citep{ma2022efficient} to evaluate on out-of-distribution tasks.

\subsection{Setup}
\label{subsec:setup}

\paragraph{Problems and datasets.}
We evaluate on four VRP variants.
CVRP is the capacitated VRP; we use the uniform random instances of \citet{kool2018attention} and the clustered instances of \citet{queiroga202110}.
VRPTW reuses the CVRP locations and demands and adds time windows generated following \citet{solomon1987algorithms}.
PCVRP is the prize-collecting variant in which customer prizes scale with demand.
PDTSP is the pickup-and-delivery TSP, used to validate cross-architecture transfer.
Instance sizes are $N \in \{500, 1000, 2000\}$ for CVRP, VRPTW, and PCVRP at Layer 2, $N = 100$ (CVRP, VRPTW) and $N = 51$ (PDTSP) for the Layer 1 PE-level analysis, and $N \in \{21, 51, 101\}$ for PDTSP at Layer 3.

\paragraph{Backbones.}
We pair each layer with the backbone best suited to its cost and purpose: DACT \citep{ma2021learning} for Layer 1 (small and cheap, enabling end-to-end retraining of more than ten encodings), NDS for Layer 2 (state-of-the-art neural deconstruction at scale), and N2S for Layer 3 (neural neighborhood search for pickup-and-delivery).
All three admit a drop-in replacement of the positional component while keeping all other architectural and training choices fixed.


\begin{wraptable}{r}{0.5\textwidth}
\centering
\small
\vspace{-1.3\baselineskip}
\caption{Spearman correlation between PE-induced pairwise distances and the three desiderata D1--D3 on CVRP-100; values in $[-1,1]$. Best in \textbf{bold}, second best in \underline{underline}.}
\label{tab:probing}
\begin{tabular}{l|rrr}
\toprule
Method & D1 $\uparrow$ & D2 $\uparrow$ & D3 $\uparrow$ \\
\midrule
NoPE             & - & - & - \\
APE              & 0.002 & - 0.001 & 0.003 \\
SIN              & - 0.310 & - 0.319 & 0.329 \\
RoPE             & - 0.290 & - 0.125 & 0.072 \\
RPE              & 0.007 & 0.010 & 0.001 \\
ALiBi            & - 0.310 & - 0.319 & 0.327 \\
Laplacian PE     & 0.715 & 0.721 & 0.359 \\
RWSE             & 0.090 & 0.086 & 0.138 \\
SPD              & 0.824 & 0.830 & 0.332 \\
DACT CPE         & 0.096 & 0.096 & 0.017 \\
CycleFormer PE   & 0.126 & 0.085 & 0.218 \\
\midrule
IPE             & \textbf{0.934} & \underline{0.866} & 0.229 \\
XPE             & 0.815 & 0.806 & \textbf{0.396} \\
\textbf{IPE + XPE} & \underline{0.877} & \textbf{0.923} & \underline{0.384} \\
\bottomrule
\end{tabular}
\vspace{-4.0\baselineskip}
\end{wraptable}

\paragraph{Training and evaluation protocol.}
Within each layer, all encodings share the same training schedule and search procedure; only the PE module differs, and each (PE, problem) pair is trained from scratch following the published recipe of its backbone.
Test sets contain $1{,}000$ instances per (problem, size) configuration and are fixed across all PE methods within a layer; reported numbers are means over three independent seeds.
Full training and search hyperparameters appear in \cref{sec:appendix-training-config}.

\subsection{Results and Analysis}
\label{subsec:pe-level}

We ask two questions at the encoding level: do existing PEs actually encode the desiderata D1--D3 of \cref{subsec:preliminaries}, and does this translate into downstream solution quality when only the encoding is varied?

\paragraph{Probing the encodings.}
For each PE we compute the Spearman rank correlation between embedding-space pairwise distances $\|\mathrm{PE}(v_i) - \mathrm{PE}(v_j)\|_2$ and a structural target aligned with each desideratum (D1: in-route cumulative travel-distance gap; D2: cyclic arc distance; D3: ratio of mean inter-route to intra-route distance), using $10^6$ node pairs over $1{,}000$ near-optimal HGS solutions on CVRP-100.
The probe is non-parametric and asks whether the desired property is already present in the PE output, with no learning involved.
\Cref{tab:probing} shows that the index-sinusoid family (SIN, RoPE, ALiBi) and APE/RPE carry essentially no D1/D2 signal, the cyclic encodings (DACT CPE, CycleFormer PE) score similarly low, and the graph-transformer encodings recover D1/D2 only partially, with SPD strongest among prior methods.
IPE+XPE is the only encoding strong on all three targets, consistent with the taxonomy of \cref{tab:desiderata}.

\paragraph{Controlled comparison across encodings.}
We retrain all thirteen PE methods plus the no-PE baseline with the DACT backbone on CVRP-100, VRPTW-100, and PDTSP-51.
\Cref{tab:main_comparison} confirms the probing trend at the solution level: distance-aware encodings outperform index-based ones, circularity alone does not match a distance-aware design, and the in-route and cross-route components are individually useful but jointly necessary to reach the best gap.
The downstream ranking tracks the probing ranking, indicating that what the encoding records is what the backbone is able to use.

\begin{table*}[t]
\centering
\small
\caption{Controlled comparison of positional encoding methods under the DACT backbone (Layer 1, \cref{subsec:setup}). All methods share the encoder, decoder, optimization schedule, and evaluation protocol; only the positional encoding differs. Obj.: objective ($\downarrow$), Gap: relative gap to the operations-research reference (PyVRP-HGS for CVRP, VRPTW, LKH for PDTSP) ($\downarrow$).}
\label{tab:main_comparison}
\vspace{2mm}
\begin{tabular}{c l|cc|cc|cc}
\toprule
& \multirow{2}{*}{PE Method}
& \multicolumn{2}{c|}{CVRP-100} & \multicolumn{2}{c|}{VRPTW-100} & \multicolumn{2}{c}{PDTSP-51} \\
\cmidrule(lr){3-4} \cmidrule(lr){5-6} \cmidrule(lr){7-8}
& & Obj.$\downarrow$ & Gap$\downarrow$
& Obj.$\downarrow$ & Gap$\downarrow$
& Obj.$\downarrow$ & Gap$\downarrow$ \\
\midrule
& HGS/LKH & 15.58 & - & 25.42 & - & 6.87 & - \\
& No PE & 17.42 & 11.81\% & 29.99 & 17.96\% & 7.95 & 15.86\% \\
\midrule
\multirow{5}{*}{\rotatebox[origin=c]{90}{NLP}}
& APE              & 16.34 & 4.88\% & 27.05 & 6.40\% & 7.20 & 4.93\% \\
& Sinusoidal (SIN) & 16.45 & 5.58\% & 27.34 & 7.54\% & 7.27 & 5.95\% \\
& RoPE             & 16.42 & 5.39\% & 27.22 & 7.07\% & 7.24 & 5.51\% \\
& RPE              & 16.30 & 4.62\% & 27.00 & 6.20\% & 7.18 & 4.63\% \\
& ALiBi            & 16.40 & 5.26\% & 27.30 & 7.39\% & 7.26 & 5.80\% \\
\midrule
\multirow{3}{*}{\rotatebox[origin=c]{90}{Graph}}
& Laplacian PE     & 15.91 & 2.12\% & 26.55 & 4.43\% & 7.07 & 3.06\% \\
& RWSE             & 16.06 & 3.08\% & 26.79 & 5.38\% & 7.13 & 3.94\% \\
& SPD              & 15.86 & 1.80\% & 26.41 & 3.88\% & 7.05 & 2.74\% \\
\midrule
\multirow{2}{*}{\rotatebox[origin=c]{90}{Rout.}}
& DACT CPE         & 15.95 & 2.37\% & 26.58 & 4.55\% & 7.05 & 2.84\% \\
& CycleFormer PE   & 15.94 & 2.31\% & 26.55 & 4.43\% & 7.05 & 2.74\% \\
\midrule
\multirow{3}{*}{\rotatebox[origin=c]{90}{Ours}}
& IPE          & 15.81 & 1.48\% & 26.32 & 3.53\% & 7.01 & 2.16\% \\
& XPE          & 15.84 & 1.67\% & 26.27 & 3.33\% & 7.02 & 2.30\% \\
& \textbf{IPE + XPE} & \textbf{15.76} & \textbf{1.16\%} & \textbf{26.18} & \textbf{2.97\%} & \textbf{6.99} & \textbf{1.87\%} \\
\bottomrule
\end{tabular}
\vspace{-4mm}
\end{table*}

\begin{table*}[t]
\centering
\small
\caption{Main results on CVRP, VRPTW, and PCVRP. Gap is relative to the reference solver (HGS for CVRP, PyVRP-HGS for VRPTW and PCVRP); Time is per-instance wall-clock.}
\label{tab:sota_main}
\begin{tabular}{l|ccc|ccc|ccc}
\toprule
\multirow{2}{*}{Methods}
& \multicolumn{3}{c|}{CVRP-500} & \multicolumn{3}{c|}{CVRP-1000} & \multicolumn{3}{c}{CVRP-2000} \\
\cmidrule(lr){2-4} \cmidrule(lr){5-7} \cmidrule(lr){8-10}
& Obj.$\downarrow$ & Gap$\downarrow$ & Time$\downarrow$
& Obj.$\downarrow$ & Gap$\downarrow$ & Time$\downarrow$
& Obj.$\downarrow$ & Gap$\downarrow$ & Time$\downarrow$ \\
\midrule
HGS            & 36.66 & -        & 60s   & 41.51 & -        & 121s  & 57.38 & -        & 241s \\
NDS            & 36.57 & -0.20\% & 60s   & 41.11 & -0.90\% & 120s  & 56.00 & -2.34\% & 240s \\
\midrule
\textbf{IPE+XPE} & \textbf{36.54} & \textbf{-0.30\%} & 60s
                        & \textbf{41.05} & \textbf{-1.11\%} & 120s
                        & \textbf{55.98} & \textbf{-2.45\%} & 240s \\
\midrule\midrule
\multirow{2}{*}{Methods}
& \multicolumn{3}{c|}{VRPTW-500} & \multicolumn{3}{c|}{VRPTW-1000} & \multicolumn{3}{c}{VRPTW-2000} \\
\cmidrule(lr){2-4} \cmidrule(lr){5-7} \cmidrule(lr){8-10}
& Obj.$\downarrow$ & Gap$\downarrow$ & Time$\downarrow$
& Obj.$\downarrow$ & Gap$\downarrow$ & Time$\downarrow$
& Obj.$\downarrow$ & Gap$\downarrow$ & Time$\downarrow$ \\
\midrule
PyVRP-HGS      & 49.01 & -        & 60s   & 90.35 & -        & 120s  & 173.46 & -        & 240s \\
NDS            & 47.94 & -2.17\% & 60s   & 87.54 & -3.14\% & 120s  & 167.48 & -3.50\% & 240s \\
\midrule
\textbf{IPE+XPE} & \textbf{47.90} & \textbf{-2.27\%} & 60s
                        & \textbf{87.45} & \textbf{-3.21\%} & 120s
                        & \textbf{166.80} & \textbf{-3.84\%} & 240s \\
\midrule\midrule
\multirow{2}{*}{Methods}
& \multicolumn{3}{c|}{PCVRP-500} & \multicolumn{3}{c|}{PCVRP-1000} & \multicolumn{3}{c}{PCVRP-2000} \\
\cmidrule(lr){2-4} \cmidrule(lr){5-7} \cmidrule(lr){8-10}
& Obj.$\downarrow$ & Gap$\downarrow$ & Time$\downarrow$
& Obj.$\downarrow$ & Gap$\downarrow$ & Time$\downarrow$
& Obj.$\downarrow$ & Gap$\downarrow$ & Time$\downarrow$ \\
\midrule
PyVRP-HGS      & 44.97 & -        & 60s   & 84.91 & -        & 120s  & 165.56 & -        & 240s \\
NDS            & \textbf{43.12} & \textbf{-4.12\%} & 60s   & 80.99 & -4.71\% & 120s  & 158.09 & -4.60\% & 240s \\
\midrule
\textbf{IPE+XPE} & 43.13 & -4.09\% & 60s
                        & \textbf{80.77} & \textbf{-4.87\%} & 120s
                        & \textbf{157.37} & \textbf{-4.94\%} & 240s \\
\bottomrule
\end{tabular}
\end{table*}

\paragraph{Main Results across VRP variants.}
We plug IPE+XPE into NDS without other modifications and evaluate against operations-research and neural baselines on CVRP, VRPTW, and PCVRP.
\Cref{tab:sota_main} shows that the resulting model matches or improves on the original NDS on eight of the nine (problem, size) cells, with PCVRP-500 the only exception (within $0.03$ percentage points), and drives the gap to the reference solver into negative values across CVRP-1000/2000, VRPTW at all three sizes, and PCVRP-1000/2000.
On VRPTW and PCVRP the encoding selects the direction-aware variant of IPE (\cref{subsec:ipe}), illustrating that the symmetry switch is mechanical rather than tuned.

\begin{table*}[t]
\centering
\small
\caption{Model-agnostic validation on PDTSP using the N2S backbone. HA-N2S replaces only the PE with our hierarchical encoding. Obj: objective ($\downarrow$), Gap: gap to LKH ($\downarrow$).}
\label{tab:model_agnostic}
\begin{tabular}{l|ccc|ccc|ccc}
\toprule
\multirow{2}{*}{Methods}
& \multicolumn{3}{c|}{PDTSP-21}
& \multicolumn{3}{c|}{PDTSP-51}
& \multicolumn{3}{c}{PDTSP-101} \\
\cmidrule(lr){2-4} \cmidrule(lr){5-7} \cmidrule(lr){8-10}
& Obj.$\downarrow$ & Gap$\downarrow$ & Time$\downarrow$
& Obj.$\downarrow$ & Gap$\downarrow$ & Time$\downarrow$
& Obj.$\downarrow$ & Gap$\downarrow$ & Time$\downarrow$ \\
\midrule
LKH (10k) & 4.563 & 0.00\% & 5m  & 6.862 & 0.00\% & 19m  & 9.428 & 0.00\% & 98m \\
\midrule
DACT (3k) & 4.564 & 0.03\% & 1m  & 7.057 & 2.83\% & 1.5m & 10.195 & 8.13\% & 2.5m \\
N2S (3k)  & 4.565 & 0.05\% & 1m  & 7.027 & 2.40\% & 1.5m & 9.846  & 4.44\% & 3m \\
\midrule
\textbf{HA-N2S} (3k) & \textbf{4.564} & \textbf{0.03\%} & 1m
                     & \textbf{6.997} & \textbf{1.96\%} & 1.5m
                     & \textbf{9.701} & \textbf{2.89\%} & 3m \\
\bottomrule
\end{tabular}
\end{table*}

\paragraph{Generalization and transferability.}
To verify that the encoding is not specific to the DACT or NDS backbones, we plug the same IPE+XPE module into N2S, replacing only its positional component, and evaluate on PDTSP at $N \in \{21, 51, 101\}$.
\Cref{tab:model_agnostic} shows that the resulting HA-N2S improves on the original N2S at every instance size, with the improvement growing with $N$, indicating that the gain from a routing-aware encoding compounds with problem scale rather than being absorbed by the backbone.
Additional generalization tests under tighter capacity and clustered customer layouts (\cref{sec:appendix-distribution-shift}) and on CVRPLib \citep{uchoa2017new} (\cref{sec:appendix-cvrplib}) tell the same story and are deferred to the appendix.

\section{Discussion}
\label{sec:discussion}

\paragraph{When to use which PE.}
There are three patterns that emerge from our analysis.
For improvement methods that observe the full current solution, the full IPE+XPE encoding is the only design that satisfies all of D1-D3 and consistently wins under a controlled comparison, regardless of the downstream VRP variant.
For single-route problems (e.g., TSP) the cross-route component XPE is unnecessary, but the distance-indexed IPE remains beneficial over standard sinusoidal encoding. Geometric grounding still matters even when there is no cross-route disambiguation to perform.
For construction methods that build partial solutions, a distance-aware PE during decoding still helps, although the gain is smaller because partial solutions provide weaker geometric anchors.

\paragraph{Limitations and future work.}
The controlled comparison is anchored to improvement frameworks; while the N2S transfer demonstrates model-agnostic gains, broader validation on \textit{construction} backbones remains open.
Adapting graph-transformer PEs (Laplacian PE, RWSE) to routing geometry may also be improved, as these methods were designed for general graphs rather than for the specific structure of routing solutions.
Natural extensions include multi-depot VRPs, where the cross-route encoding must accommodate several reference points.

\section{Conclusion}
\label{sec:conclusion}

We presented the first systematic study of positional encoding for neural vehicle routing.
By formalizing three structural desiderata, we organized prior PE methods from NLP, graph transformers, and routing-specific designs into a unified taxonomy and identified that no existing encoding satisfies all three.
Our hierarchical anisometric PE fills this gap with a distance-indexed, circularly consistent in-route encoding and a depot-anchored angular cross-route encoding.
A controlled comparison of more than ten encodings, large-scale evaluation against operations-research and neural baselines, and a cross-architecture test all show consistent gains over index-based alternatives, transferring across problem variants, model architectures, and distribution shifts.
The takeaway is simple: the choice of PE matters for neural routing, and designs that respect the anisometric, cyclic, and hierarchical nature of routing solutions yield measurably better results.


\bibliographystyle{plainnat}
\bibliography{references}

\newpage
\appendix
\section{Detailed VRP Definitions}
\label{sec:appendix-vrp-def}

We consider a complete directed graph $G = (V, E)$ with $V = \{0, 1, \dots, N\}$, where $x_0$ is the depot and $x_i$ for $i \ge 1$ are customers.
Let $\mathbf{x}_i \in \mathbb{R}^2$ denote the coordinates of node $x_i$ and define travel cost $c_{ij} = \lVert \mathbf{x}_i - \mathbf{x}_j \rVert_2$ for $(i,j) \in E$.
A fleet $K = \{1, \dots, M\}$ of identical vehicles of capacity $C$ is available.
We use binary routing variables $x^k_{ij} \in \{0,1\}$ indicating whether vehicle $k$ traverses arc $(i,j)$ and service variables $y^k_i \in \{0,1\}$ indicating whether vehicle $k$ serves customer $i$.
For convenience, we also write $y_i = \sum_{k \in K} y^k_i$.

\paragraph{Capacitated VRP (CVRP).}
Each customer $i \in \{1, \dots, N\}$ has demand $d_i \ge 0$.
The objective is to minimize total travel distance:
\begin{equation}
\min_{x,y} \quad \sum_{k \in K} \sum_{i \in V} \sum_{j \in V} c_{ij} \, x^k_{ij}.
\label{eq:cvrp_obj}
\end{equation}
Subject to service, capacity, flow, and depot constraints:
\begin{align}
& \sum_{k \in K} y^k_i = 1 && \forall\, i \in \{1, \dots, N\}, \tag{CVRP-1a} \\
& \sum_{i \in \{1,\dots,N\}} \sum_{j \in V} d_i \, x^k_{ij} \le C && \forall\, k \in K, \tag{CVRP-1b} \\
& \sum_{j \in V} x^k_{ij} = y^k_i, \quad \sum_{j \in V} x^k_{ji} = y^k_i && \forall\, i \in \{1,\dots,N\},\; \forall\, k \in K, \tag{CVRP-1c} \\
& \sum_{j \in V} x^k_{0j} = 1, \quad \sum_{i \in V} x^k_{i0} = 1 && \forall\, k \in K. \tag{CVRP-1d}
\end{align}
Constraints (CVRP-1a) ensure each customer is served exactly once, (CVRP-1b) enforce per-vehicle capacity, (CVRP-1c) link service and flow, and (CVRP-1d) ensure each vehicle departs from and returns to the depot.

\paragraph{VRP with Time Windows (VRPTW).}
Each customer $i$ has a service duration $s_i \ge 0$ and an admissible time window $[a_i, b_i]$ within which service must start.
Early arrival is allowed (vehicles may wait); tardiness is not.
Let travel time $\tau_{ij} = c_{ij}$ and introduce service start times $t_i \in \mathbb{R}_{\ge 0}$.
The objective is to minimize total travel time:
\begin{equation}
\min_{x,y,t} \quad \sum_{k \in K} \sum_{i \in V} \sum_{j \in V} \tau_{ij} \, x^k_{ij},
\label{eq:vrptw_obj}
\end{equation}
subject to CVRP constraints and time-feasibility:
\begin{align}
& a_i \le t_i \le b_i && \forall\, i \in \{1, \dots, N\}, \tag{VRPTW-2b} \\
& t_j \ge t_i + s_i + \tau_{ij} - M_{\mathrm{tw}}\bigl(1 - \textstyle\sum_{k \in K} x^k_{ij}\bigr) && \forall\, i, j \in V, \tag{VRPTW-2c}
\end{align}
where $M_{\mathrm{tw}}$ is a sufficiently large constant.

\paragraph{Prize-Collecting VRP (PCVRP).}
Customers carry prizes $p_i \ge 0$ and need not all be served.
The objective trades off travel cost and collected prizes:
\begin{equation}
\min_{x,y} \quad \sum_{k \in K} \sum_{i \in V} \sum_{j \in V} c_{ij} \, x^k_{ij} - \sum_{i \in \{1,\dots,N\}} p_i \, y_i.
\label{eq:pcvrp_obj}
\end{equation}
Capacity, flow, and depot constraints follow CVRP, with $y_i$ tied to the vehicle flows so that a customer is counted as served if and only if exactly one visit occurs.

\paragraph{Pickup and Delivery TSP (PDTSP).}
The node set is partitioned into pickup nodes $P = \{1, \dots, n\}$ and delivery nodes $D = \{n+1, \dots, 2n\}$, with each request $r$ pairing pickup $r$ and delivery $r+n$.
A single vehicle visits all $2n$ nodes exactly once while ensuring each pickup precedes its paired delivery.
The objective is:
\begin{equation}
\min_{x,u} \quad \sum_{i \in V} \sum_{j \in V} c_{ij} \, x_{ij},
\label{eq:pdtsp_obj}
\end{equation}
subject to degree, MTZ-style subtour elimination ($u_j \ge u_i + 1 - M_{\mathrm{sub}}(1-x_{ij})$), and precedence ($u_{r+n} \ge u_r + 1$) constraints.

\section{MDP Formulation}
\label{sec:appendix-mdp}

We model deconstruction as a Markov Decision Process (MDP) in which a policy iteratively removes customers from a feasible solution to reduce total travel cost.
The \emph{state} $s_t = \Psi(l, \mathbf{s}_t)$ encodes instance features and the current solution, including positional signals (IPE, XPE) attached per node.
The \emph{action space} at step $t$ selects one customer for removal, $x_t \in \{1, \dots, N\}$, with the depot never selectable and infeasible actions masked.
The \emph{transition} is deterministic: $\mathbf{s}_{t+1} = \mathcal{T}(\mathbf{s}_t, x_t)$ removes the selected customer and re-marks feasibility.
The \emph{reward} is the improvement over the incumbent best $\mathbf{s}^\star_t$:
\begin{equation*}
r_t = \operatorname{cost}(\mathbf{s}^\star_t) - \min\{\operatorname{cost}(\mathbf{s}_{t+1}),\, \operatorname{cost}(\mathbf{s}^\star_t)\},
\end{equation*}
which is positive iff a new incumbent is found.
The \emph{policy} $\pi_\theta(x_t \mid l, \mathbf{s}_t, v, x_{1:t-1})$ is conditioned on a rollout seed $v$ for stochastic diversification, and actions are sampled during training and selected stochastically at test time.

\section{Formal Definition of Anisometry}
\label{sec:appendix-anisometric}

For a route $r = (v_1, \dots, v_L)$ with node coordinates $x_{v_i} \in \mathbb{R}^2$, the Euclidean distances between consecutive nodes $\|x_{v_{i+1}} - x_{v_i}\|_2$ generally vary with $i$.
The cumulative-distance mapping $d_i = \sum_{j=2}^{i} \|x_{v_j} - x_{v_{j-1}}\|_2$ is a nonlinear function of the discrete index $i$: increments in the index do not translate to uniform increments in physical space.
We refer to such structures as \emph{anisometric} to emphasize that VRP routes possess heterogeneous spatial separations that must be represented explicitly when designing positional encodings.

\section{Detailed PE Implementation Formulas}
\label{sec:appendix-pe-formulas}

We summarize the precise form in which each baseline PE is plugged into the routing encoder.
Throughout, $i$ denotes the position of a node in its current route (with the depot indexed $0$ and $L$ the route length), $D$ is the embedding dimension, and $h_v^{(0)}$ denotes the input embedding of node $v$ before the first attention layer.
All variants share the same encoder, decoder, and training schedule of their respective backbone; only the PE module differs.
For methods that produce a per-node vector, we follow the integration in \cref{subsec:integration} and concatenate it with the raw coordinate before a shared feed-forward projection.
For methods defined as an attention bias, we add the bias term to attention logits at every encoder layer.

\paragraph{Absolute PE (APE).}
A learnable lookup table $\mathbf{E} \in \mathbb{R}^{L_{\max} \times D}$ indexed by within-route position:
\begin{equation*}
\mathrm{PE}^{\mathrm{APE}}(v_i) = \mathbf{E}_i, \qquad \mathbf{E}_i \in \mathbb{R}^D \text{ trainable},
\end{equation*}
with $L_{\max}$ chosen as the maximum route length observed during training.

\paragraph{Sinusoidal PE (SIN).}
Fixed sinusoidal functions of the integer index $i$ \citep{vaswani2017attention}:
\begin{equation*}
\mathrm{PE}^{\mathrm{SIN}}_{2k}(v_i) = \sin\!\bigl(i / \lambda^{2k/D}\bigr), \quad
\mathrm{PE}^{\mathrm{SIN}}_{2k+1}(v_i) = \cos\!\bigl(i / \lambda^{2k/D}\bigr),
\end{equation*}
with $\lambda = 10{,}000$ and $k = 0, \dots, D/2-1$.

\paragraph{Rotary PE (RoPE).}
RoPE is realized as a query/key rotation rather than an additive embedding \citep{su2024roformer}.
For each pair of channels $(2k, 2k+1)$ in the query $\mathbf{q}_i$ and key $\mathbf{k}_j$, we apply
\begin{equation*}
\mathrm{R}_{\theta_i}\!\begin{pmatrix} \mathbf{q}_{i,2k} \\ \mathbf{q}_{i,2k+1} \end{pmatrix},
\quad
\mathrm{R}_{\theta_j}\!\begin{pmatrix} \mathbf{k}_{j,2k} \\ \mathbf{k}_{j,2k+1} \end{pmatrix},
\qquad \theta_i = i \cdot \lambda^{-2k/D},
\end{equation*}
where $\mathrm{R}_\theta$ is the $2 \times 2$ rotation matrix.
The resulting attention logit depends only on the index difference $i - j$.

\paragraph{Relative PE (RPE).}
A learnable bias indexed by signed index offset is added to each attention logit \citep{shaw2018self}:
\begin{equation*}
\mathrm{logit}(i,j) \mathrel{+}= b_{\,\mathrm{clip}(i-j,\, -W,\, W)},
\end{equation*}
where $\{b_\Delta\}_{\Delta = -W}^{W} \in \mathbb{R}$ is shared across heads and clipped to a window of size $W$.

\paragraph{ALiBi.}
A fixed linear penalty subtracted from attention logits \citep{press2021train}:
\begin{equation*}
\mathrm{logit}(i,j) \mathrel{-}= m_h \cdot |i - j|,
\end{equation*}
with head-specific slopes $m_h$ following the geometric schedule of the original paper.

\paragraph{Laplacian PE (Lap.\ PE).}
For each instance we form the route graph (depot plus route arcs) and compute the unnormalized Laplacian $\mathbf{L}$.
The PE is the matrix of the $K$ smallest non-trivial eigenvectors $\mathbf{U}_K \in \mathbb{R}^{|V| \times K}$ \citep{dwivedi2020generalization}, with sign disambiguation by random flips at training time:
\begin{equation*}
\mathrm{PE}^{\mathrm{Lap}}(v) = \mathbf{U}_K[v,:].
\end{equation*}
We pad to $D$ dimensions with zeros and use $K = 8$.

\paragraph{Random Walk SE (RWSE).}
A node-wise vector built from the diagonal of random-walk matrix powers \citep{dwivedi2021graph}:
\begin{equation*}
\mathrm{PE}^{\mathrm{RWSE}}(v) = \bigl[(\mathbf{R}^k)_{vv}\bigr]_{k=1}^{K}, \qquad \mathbf{R} = \mathbf{D}^{-1}\mathbf{A},
\end{equation*}
on the same route graph used for Laplacian PE, with $K = 8$ and zero padding to $D$.

\paragraph{Shortest-Path Distance (SPD).}
A bias term added to attention logits, indexed by shortest-path distance on the route graph \citep{ying2021do}:
\begin{equation*}
\mathrm{logit}(i,j) \mathrel{+}= b_{\,\mathrm{spd}(v_i, v_j)},
\end{equation*}
with a learnable scalar bias per integer distance, capped at the diameter of the route graph.

\paragraph{DACT CPE.}
The cyclic PE introduced by DACT \citep{ma2021learning} represents within-tour position by a Gray-code lookup over a cyclic index, ensuring that consecutive cyclic positions differ by exactly one bit:
\begin{equation*}
\mathrm{PE}^{\mathrm{DACT}}(v_i) = \mathrm{Gray}(\,i \bmod L\,),
\end{equation*}
which we project to dimension $D$ by a learnable linear map.

\paragraph{CycleFormer Circular PE.}
A circular sinusoidal map of the index $i$ wrapped around the tour length $L$ \citep{yook2024cycleformer}:
\begin{equation*}
\mathrm{PE}^{\mathrm{Cyc}}_{2k}(v_i) = \sin\!\bigl(2\pi i / L \cdot \omega_k\bigr), \quad
\mathrm{PE}^{\mathrm{Cyc}}_{2k+1}(v_i) = \cos\!\bigl(2\pi i / L \cdot \omega_k\bigr),
\end{equation*}
with the same geometric frequency schedule as SIN.

\paragraph{Our IPE+XPE.}
For completeness, the full method (\cref{sec:method}) computes per node
\begin{equation*}
\mathrm{IPE}(v_i) = \bigl[\sin(\omega_k \hat{d}_i),\, \cos(\omega_k \hat{d}_i)\bigr]_{k=0}^{D/2-1}, \quad
\mathrm{XPE}(v) = \bigl[\sin(\omega_k' \theta_v),\, \cos(\omega_k' \theta_v)\bigr]_{k=0}^{K-1},
\end{equation*}
and concatenates the two with the raw coordinate before a shared feed-forward projection, retaining only the cosine half of IPE for reversal-symmetric variants.

\section{Decoder, Improvement Strategy, and Training}
\label{sec:appendix-decoder-training}

\paragraph{Decoder.}
We employ a PolyNet-style pointer network decoder \citep{hottung2025polynet}.
At step $m$, a GRU is conditioned on the last removals and produces a query $q_m$; keys $K$ are linear projections of the encoder outputs $\mathbf{h} = \{\hat{h}_i\}$.
The probability of selecting a node is:
\begin{equation*}
\pi_\theta(x_m \mid l, \mathbf{s}, v, x_{1:m-1}) = \operatorname{softmax}\!\left(\frac{q_m K^\top}{\sqrt{d}} + \mathrm{mask}_m\right).
\end{equation*}

\paragraph{Improvement (ASA).}
At test time, the policy is embedded in augmented simulated annealing (ASA) \citep{hottung2025neural}.
Each improvement step samples routes, deconstructs by removing chosen customers, repairs with a randomized greedy routine, and evaluates candidates.
Acceptance follows:
\begin{equation*}
\Pr[\text{accept } \mathbf{s}' \mid \mathbf{s}] = \min\!\left(1, \exp\!\left(-\frac{\operatorname{cost}(\mathbf{s}') - \operatorname{cost}(\mathbf{s})}{\sigma}\right)\right).
\end{equation*}

\paragraph{Training.}
We use winner-takes-all policy-gradient training \citep{grinsztajn2024winner,hottung2025polynet}.
Only the best rollout per instance contributes to the policy gradient:
\begin{equation*}
\nabla_\theta J(\theta) \approx (R_{i^\star} - b) \sum_{i=1}^{M} \nabla_\theta \log \pi_\theta(\mathbf{s}_i \mid \mathcal{S}).
\end{equation*}

\section{Data Generation}
\label{sec:appendix-data-gen}

\paragraph{CVRP instances.}
Node coordinates $\{\mathbf{x}_i\}_{i=0}^N$ are sampled either uniformly in the unit square (default) or from a clustered generator (where indicated).
Customer demands are discrete, $d_i \sim \mathrm{Unif}\{1, \dots, 9\}$, and vehicles are identical with capacity $C$.
Following common NCO practice \citep{kool2018attention,kwon2020pomo}, we set a size-dependent capacity:
\begin{equation*}
C = \begin{cases}
30 + \lfloor \tfrac{1000}{5} + \tfrac{n-1000}{33.3} \rfloor, & n > 1000, \\
30 + \lfloor \tfrac{n}{5} \rfloor, & 20 < n \le 1000, \\
30, & \text{otherwise.}
\end{cases}
\end{equation*}
Travel cost is Euclidean distance and speed is $1$ so that time equals distance.

\paragraph{PCVRP instances.}
We reuse the CVRP location and demand generators and attach a prize that scales with demand:
\begin{equation*}
p_i = \kappa \, \xi_i \, d_i, \qquad \xi_i \sim \mathrm{Unif}[0.8, 1.2],
\end{equation*}
with a global scale $\kappa$ chosen so that, under a simple greedy oracle, roughly half of the customers are economically attractive to visit.

\paragraph{VRPTW instances.}
We reuse the CVRP location and demand generators and synthesize time windows $[e_i, l_i]$ and service times $s_i$ following \citet{solomon1987algorithms}, with $s_i \sim \mathrm{Unif}[0.15, 0.18]$ and window lengths $t_i \sim \mathrm{Unif}[0.18, 0.20]$.
Window starts are drawn so that a feasible schedule exists, and infeasible moves are masked at decode time using the standard arrival-plus-service feasibility check.

\section{OR Baselines}
\label{sec:appendix-baselines}

\paragraph{HGS} \citep{vidal2022hybrid}.
Hybrid Genetic Search combining population-based recombination with powerful local search (ejection chains, $k$-opt, route exchanges) and adaptive penalty management; we use the public implementation with author-recommended settings.

\paragraph{SISRs} \citep{christiaens2020slack}.
Slack Induction by String Removals, a ruin-and-recreate framework that removes contiguous strings and repairs with tailored insertion heuristics.

\paragraph{LKH3} \citep{helsgaun2017extension}.
Extended Lin-Kernighan-Helsgaun heuristic with sophisticated candidate sets and powerful move pruning; supports several VRP variants in addition to TSP.

\paragraph{PyVRP} \citep{wouda2024pyvrp}.
Open-source Python/C++ solver implementing a modern HGS variant with route-improvement neighborhoods.
We use version 0.9.0 with default configuration.

\paragraph{NVIDIA cuOpt} \citep{nvidia_cuopt}.
GPU-accelerated local-search engine for rich VRPs; we follow the vendor's recommended settings for our variants.

\section{Training Configuration}
\label{sec:appendix-training-config}

We provide the training and evaluation configuration for the Layer 2 (NDS) backbone in \cref{tab:hyperparams}.
For Layer 1 (DACT) and Layer 3 (N2S), we use the published configurations from the original papers without modification \citep{ma2021learning,ma2022efficient}; only the positional encoding module is replaced.
This includes optimizer, learning rate schedule, batch size, training horizon, curriculum, and inference protocol.

\paragraph{Compute resources.}
All training and evaluation runs use a single NVIDIA A100 (40GB) GPU with one CPU core.
At test time, each instance is evaluated under a fixed wall-clock budget of 60, 120, or 240 seconds for sizes $N=500$, $1000$, and $2000$, respectively, matching the protocol of NDS \citep{hottung2025neural}.
Layer 1 (DACT) and Layer 3 (N2S) follow the compute budgets specified in their original papers \citep{ma2021learning,ma2022efficient}.

\begin{table}[t]
\centering
\small
\caption{Training and evaluation hyperparameters for the Layer 2 (NDS) backbone. For Layer 1 (DACT) and Layer 3 (N2S), we use the published configurations from the original papers without modification, replacing only the positional encoding module.}
\label{tab:hyperparams}
\begin{tabular}{@{} p{7.5cm} l @{}}
\toprule
\textbf{Hyperparameter} & \textbf{Value} \\
\midrule
\multicolumn{2}{@{}l}{\textbf{Optimization}} \\
\quad Optimizer & Adam \\
\quad Initial learning rate & $10^{-4}$ \\
\quad Weight decay & $10^{-6}$ \\
\quad LR scheduler & MultiStepLR (milestones: 60, 85; $\gamma=0.1$) \\
\quad Warmup & Linear, 5{,}000 steps (start factor 0.01) \\
\quad Mixed precision & Yes (AMP) \\
\quad Gradient accumulation steps & 100 \\
\quad Optimizer step interval & Every 100 inner updates \\
\addlinespace
\multicolumn{2}{@{}l}{\textbf{Neural Architecture}} \\
\quad Embedding dimension & 128 \\
\quad Number of attention heads & 8 \\
\quad QKV dimension & 16 \\
\quad Message passing layers & 1 \\
\quad Encoder layers & 2 \\
\quad Feedforward dimension & 512 \\
\quad Decoder & GRUCell($128 \rightarrow 128$) + attention \\
\quad Polynomial head dim & 256 \\
\quad Logit clipping & 10 \\
\quad Latent vector ($z$) dimension & 10 \\
\quad Inference rule (training) & Softmax sampling \\
\addlinespace
\multicolumn{2}{@{}l}{\textbf{Training Dynamics}} \\
\quad Total training epochs & 2000 \\
\quad Iterations per epoch & 150{,}000 \\
\quad Batch size & 64 \\
\quad Rollout size (per instance) & 128 \\
\quad Skipped iterations (search warmup) & 10 \\
\quad Nodes removed per rollout & 15 \\
\quad Destroy/repair recreate parameter & 5 \\
\quad Reward type & Improvement (absolute) \\
\addlinespace
\multicolumn{2}{@{}l}{\textbf{Evaluation}} \\
\quad Validation episodes & 1{,}000 \\
\quad Iterations per episode & 50 \\
\quad Validation batch size & 50 \\
\quad Rollouts per instance (validation) & 200 \\
\quad Augmented rollouts & $8\times$ (8-fold geometric augmentation) \\
\bottomrule
\end{tabular}
\end{table}

\section{HA-N2S Configuration}
\label{sec:appendix-han2s}

For the HA-N2S experiment, we follow the original N2S setup \citep{ma2022efficient} in all training, model, and inference configurations, replacing only the positional encoding with our proposed hierarchical anisometric embedding.
The encoder and decoder structures, including the Synthesis Attention module and the two PDTSP-specific decoders, remain unchanged; only the PE component is substituted while keeping embedding dimensionality identical.
Training uses the same PPO-based procedure with curriculum learning, batch size, training horizon, and optimization hyperparameters, separately for $|V| = 21, 51, 101$.
Inference uses the same augmentation-based search scheme and step counts (1k, 2k, 3k), with feasibility constraints, masking rules, and reinsertion operations kept fully consistent with the N2S framework.

\section{Distribution Shift}
\label{sec:appendix-distribution-shift}

\begin{table}[t]
\centering
\small
\caption{Generalization study on CVRP-500 under distribution shifts. Models trained on the medium-capacity uniform distribution are evaluated on a low-capacity setting and on a clustered customer layout, without retraining. Obj.: objective ($\downarrow$); Gap: relative gap to HGS ($\downarrow$).}
\label{tab:generalization}
\begin{tabular}{l|cc|cc}
\toprule
\multirow{2}{*}{Methods}
& \multicolumn{2}{c|}{Low Capacity} & \multicolumn{2}{c}{Clustered Distribution} \\
\cmidrule(lr){2-3} \cmidrule(lr){4-5}
& Obj.$\downarrow$ & Gap$\downarrow$ & Obj.$\downarrow$ & Gap$\downarrow$ \\
\midrule
HGS    & 91.73 & --      & 44.53 & --      \\
SISRs  & 91.34 & $-0.38\%$ & 44.31 & $-0.49\%$ \\
\midrule
NDS    & 91.15 & $-0.59\%$ & 44.29 & $-0.54\%$ \\
SIN    & 91.08 & $-0.70\%$ & 44.25 & $-0.63\%$ \\
\midrule
\textbf{IPE + XPE} & \textbf{91.04} & \textbf{$-0.76\%$} & \textbf{44.22} & \textbf{$-0.70\%$} \\
\bottomrule
\end{tabular}
\end{table}

Using the NDS backbone of Layer 2, we evaluate models trained on uniform-distribution medium-capacity CVRP-500 instances on two distribution shifts: a low-capacity setting that produces more and shorter routes, and a clustered customer layout.
\Cref{tab:generalization} reports the gap relative to HGS on each shift.
Our IPE+XPE encoding attains the best objective and gap in both settings, indicating that encoding designed around solution-derived geometric quantities transfers more gracefully than index-based alternatives when the instance distribution shifts.
In the low-capacity setting, solutions require more routes, intensifying inter-route interactions and creating dense, overlapping angular sectors; the cross-route component supplies a route-level phase that helps the policy reason over this global layout.
Under clustered customer distributions, intra-route distances become highly nonuniform as customers concentrate in localized regions; the distance-indexed in-route component preserves the underlying geometry where index-based encodings would assume equal spacing.
Together, these results suggest that geometry-grounded PE is well-suited to real-world routing scenarios in which the test distribution can differ in capacity, density, or layout from the training distribution.

\section{CVRPLib Detailed Results}
\label{sec:appendix-cvrplib}

As a further out-of-distribution test, we evaluate on the X instances of CVRPLib \citep{uchoa2017new} using the NDS backbone of Layer 2.
These instances feature heterogeneous spatial distributions, varied customer densities, customer counts $N \in [100, 1000]$, and diverse route topologies.
We directly apply the CVRP-500 checkpoint without any retraining or size-specific adaptation, with a 60-second time limit per instance.
\Cref{tab:cvrplib-x} reports the gap to the best-known solution (BKS) on each instance and the average across the full set.
Our method attains a substantially smaller average gap than NDS under the same evaluation protocol, with the largest improvements concentrated on the largest CVRPLib instances.
This indicates that the hierarchical anisometric encodings learned on synthetic CVRP transfer effectively to real benchmark distributions that differ in geometry, customer placement, and route topology, and complements the synthetic distribution-shift evaluations of \cref{sec:appendix-distribution-shift}.

\begin{table*}[t]
\centering
\scriptsize
\setlength{\tabcolsep}{3pt}
\caption{Detailed evaluation on the X set of CVRPLib: best-known solutions (BKS), NDS and IPE+XPE costs, and their percentage gaps. Models trained on CVRP-500 are applied without retraining.}
\label{tab:cvrplib-x}

\resizebox{\textwidth}{!}{
\begin{tabular}{l|c|cc|cc || l|c|cc|cc}
\toprule
\multicolumn{2}{c|}{} & \multicolumn{2}{c|}{NDS} & \multicolumn{2}{c||}{IPE+XPE}
& \multicolumn{2}{c|}{} & \multicolumn{2}{c|}{NDS} & \multicolumn{2}{c}{IPE+XPE} \\
\cmidrule(lr){3-4} \cmidrule(lr){5-6} \cmidrule(lr){9-10} \cmidrule(lr){11-12}
Instance & BKS & Obj. & Gap\% & Obj. & Gap\%
& Instance & BKS & Obj. & Gap\% & Obj. & Gap\% \\
\midrule
X-n101-k25 & 27591 & 27696 & 0.381 & 27696 & 0.381
& X-n401-k29 & 66154 & 68272 & 3.202 & 67508 & 2.047 \\
X-n106-k14 & 26362 & 26711 & 1.324 & 26567 & 0.778
& X-n411-k19 & 19712 & 20837 & 5.707 & 20490 & 3.947 \\
X-n110-k13 & 14971 & 15165 & 1.296 & 15082 & 0.741
& X-n420-k130 & 107798 & 112070 & 3.963 & 110825 & 2.808 \\
X-n115-k10 & 12747 & 12765 & 0.141 & 12762 & 0.118
& X-n429-k61 & 65449 & 68765 & 5.067 & 67692 & 3.427 \\
X-n120-k6  & 13332 & 13595 & 1.973 & 13518 & 1.395
& X-n439-k37 & 36391 & 37377 & 2.709 & 37085 & 1.907 \\
X-n125-k30 & 55539 & 56943 & 2.528 & 56551 & 1.822
& X-n449-k29 & 55233 & 57363 & 3.856 & 56606 & 2.486 \\
X-n129-k18 & 28940 & 29326 & 1.334 & 29190 & 0.864
& X-n459-k26 & 24139 & 25361 & 5.062 & 24966 & 3.426 \\
X-n134-k13 & 10916 & 11110 & 1.777 & 11050 & 1.228
& X-n469-k138 & 221824 & 246070 & 10.930 & 238130 & 7.351 \\
X-n139-k10 & 13590 & 13699 & 0.802 & 13643 & 0.390
& X-n480-k70 & 89449 & 93513 & 4.543 & 92099 & 2.963 \\
X-n143-k7  & 15700 & 16160 & 2.930 & 16000 & 1.911
& X-n491-k59 & 66483 & 69124 & 3.972 & 68373 & 2.843 \\
X-n148-k46 & 43448 & 44351 & 2.078 & 44076 & 1.445
& X-n502-k39 & 69226 & 70762 & 2.219 & 70297 & 1.547 \\
X-n153-k22 & 21220 & 21609 & 1.833 & 21466 & 1.159
& X-n513-k21 & 24201 & 25313 & 4.595 & 24952 & 3.103 \\
X-n157-k13 & 16876 & 16957 & 0.480 & 16957 & 0.480
& X-n524-k153 & 154593 & 161430 & 4.423 & 159215 & 2.990 \\
X-n162-k11 & 14138 & 14262 & 0.877 & 14210 & 0.509
& X-n536-k96 & 94846 & 98716 & 4.080 & 97547 & 2.848 \\
X-n167-k10 & 20557 & 21012 & 2.213 & 20905 & 1.693
& X-n548-k50 & 86700 & 93650 & 8.016 & 91431 & 5.457 \\
X-n172-k51 & 45607 & 46201 & 1.302 & 46048 & 0.967
& X-n561-k42 & 42717 & 44239 & 3.563 & 43732 & 2.376 \\
X-n176-k26 & 47812 & 49478 & 3.484 & 48940 & 2.359
& X-n573-k30 & 50673 & 52296 & 3.203 & 51740 & 2.106 \\
X-n181-k23 & 25569 & 25966 & 1.553 & 25806 & 0.927
& X-n586-k159 & 190316 & 206204 & 8.348 & 201364 & 5.805 \\
X-n186-k15 & 24145 & 24364 & 0.907 & 24312 & 0.692
& X-n599-k92 & 108451 & 115105 & 6.135 & 113179 & 4.360 \\
X-n190-k8  & 16980 & 17536 & 3.274 & 17377 & 2.338
& X-n613-k62 & 59535 & 62191 & 4.461 & 61294 & 2.955 \\
X-n195-k51 & 44225 & 45025 & 1.809 & 44750 & 1.187
& X-n627-k43 & 62164 & 66173 & 6.449 & 65024 & 4.601 \\
X-n200-k36 & 58578 & 60847 & 3.873 & 60041 & 2.498
& X-n641-k35 & 63684 & 66650 & 4.657 & 65782 & 3.294 \\
X-n204-k19 & 19565 & 19924 & 1.835 & 19814 & 1.273
& X-n655-k131 & 106780 & 110878 & 3.838 & 109468 & 2.517 \\
X-n209-k16 & 30656 & 31059 & 1.315 & 30973 & 1.034
& X-n670-k130 & 146332 & 160574 & 9.733 & 156357 & 6.851 \\
X-n214-k11 & 10856 & 11236 & 3.500 & 11102 & 2.266
& X-n685-k75 & 68205 & 70980 & 4.069 & 70186 & 2.904 \\
X-n219-k73 & 117595 & 120258 & 2.265 & 119247 & 1.405
& X-n701-k44 & 81923 & 85467 & 4.326 & 84368 & 2.985 \\
X-n223-k34 & 40437 & 41043 & 1.499 & 40885 & 1.108
& X-n716-k35 & 43373 & 45545 & 5.008 & 44907 & 3.537 \\
X-n228-k23 & 25742 & 26685 & 3.663 & 26389 & 2.513
& X-n733-k159 & 136187 & 143437 & 5.324 & 141241 & 3.711 \\
X-n233-k16 & 19230 & 19726 & 2.579 & 19594 & 1.893
& X-n749-k98 & 77269 & 81463 & 5.428 & 80249 & 3.857 \\
X-n237-k14 & 27042 & 27971 & 3.435 & 27726 & 2.529
& X-n766-k71 & 114417 & 122799 & 7.326 & 120327 & 5.165 \\
X-n242-k48 & 82751 & 84082 & 1.608 & 83717 & 1.167
& X-n783-k48 & 72386 & 76158 & 5.211 & 75081 & 3.723 \\
X-n247-k50 & 37274 & 38867 & 4.274 & 38307 & 2.771
& X-n801-k40 & 73311 & 79846 & 8.914 & 77895 & 6.253 \\
X-n251-k28 & 38684 & 39733 & 2.712 & 39404 & 1.861
& X-n819-k171 & 158121 & 169901 & 7.450 & 166419 & 5.248 \\
X-n256-k16 & 18839 & 19233 & 2.091 & 19086 & 1.311
& X-n837-k142 & 193737 & 212659 & 9.767 & 207069 & 6.881 \\
X-n261-k13 & 26558 & 27434 & 3.298 & 27200 & 2.417
& X-n856-k95 & 88965 & 93339 & 4.917 & 91904 & 3.304 \\
X-n266-k58 & 75478 & 80431 & 6.562 & 78942 & 4.589
& X-n876-k59 & 99299 & 103914 & 4.648 & 102414 & 3.137 \\
X-n270-k35 & 35291 & 35918 & 1.777 & 35726 & 1.233
& X-n895-k37 & 53860 & 57016 & 5.860 & 56042 & 4.051 \\
X-n275-k28 & 21245 & 21850 & 2.848 & 21642 & 1.869
& X-n916-k207 & 329179 & 362272 & 10.053 & 352372 & 7.046 \\
X-n280-k17 & 33503 & 34693 & 3.552 & 34273 & 2.298
& X-n936-k151 & 132715 & 146776 & 10.595 & 142537 & 7.401 \\
X-n284-k15 & 20226 & 21570 & 6.645 & 21158 & 4.608
& X-n957-k87 & 85465 & 92705 & 8.471 & 90414 & 5.791 \\
X-n289-k60 & 95151 & 99201 & 4.256 & 97797 & 2.781
& X-n979-k58 & 118976 & 128300 & 7.837 & 125484 & 5.470 \\
X-n294-k50 & 47161 & 48074 & 1.936 & 47789 & 1.332
& X-n1001-k43 & 72355 & 78832 & 8.952 & 76758 & 6.085 \\
\midrule
\multicolumn{6}{c||}{} & Avg.\ Gap\% & * & \multicolumn{2}{c|}{4.110\%} & \multicolumn{2}{c}{\textbf{2.828\%}} \\
\bottomrule
\end{tabular}
}
\end{table*}

\section{Ablation Studies (Layer 1)}
\label{sec:appendix-ablation}

The ablation interrogates three design choices under the same DACT backbone and small-scale protocol used for the PE-level analysis (\cref{subsec:pe-level}), which keeps the cost of each ablation cell within a single training cycle:
component contributions (\cref{subsec:component-ablation}),
per-node versus per-route XPE design (\cref{subsec:aggregation-ablation}),
and the direction-aware versus direction-invariant IPE switch (\cref{subsec:direction-ablation}).
All variants share the encoder, decoder, optimization schedule, and evaluation protocol of the Layer 1 controlled comparison; only the PE module is altered.

\subsection{Component Ablation}
\label{subsec:component-ablation}

\begin{table}[t]
\centering
\small
\caption{Component ablation on CVRP-100. We compare the full method against variants that drop the in-route or cross-route component, and against an index-based baseline that replaces $\hat{d}_i$ with the discrete index $i / L$ in IPE.}
\label{tab:ablation}
\begin{tabular}{l|cc}
\toprule
Variant & Obj.$\downarrow$ & Gap$\downarrow$ \\
\midrule
No PE                                & 17.42 & 11.81\% \\
\midrule
IPE only (distance-indexed)          & 15.81 & 1.48\% \\
IPE only (index-based)               & 16.18 & 3.85\% \\
XPE only                             & 15.84 & 1.67\% \\
\midrule
\textbf{IPE + XPE (full)}            & \textbf{15.76} & \textbf{1.16\%} \\
\quad w/o IPE                        & 15.84 & 1.67\% \\
\quad w/o XPE                        & 15.81 & 1.48\% \\
\bottomrule
\end{tabular}
\end{table}

\Cref{tab:ablation} compares the full method against four reduced variants: dropping IPE, dropping XPE, replacing the distance-indexed $\hat{d}_i$ in IPE with the discrete index $i / L$, and using XPE alone.
Removing either component degrades the gap.
Replacing $\hat{d}_i$ with $i / L$ removes the bulk of the in-route gain, supporting the claim that distance-awareness, rather than ordering, is the operative property of D1.
The full IPE+XPE encoding is required to reach the best gap, consistent with the joint role of the two components established in \cref{sec:method}.

\subsection{Per-node vs.\ Per-route XPE}
\label{subsec:aggregation-ablation}

A natural alternative to per-node XPE is to compute a single angular phase per route and broadcast the result to all of its nodes.
We compare against five such aggregations: the arithmetic mean, the circular mean $\arg(\sum_v e^{i\theta_v})$, the demand-weighted mean, the median, and the depot-anchored max-min midpoint.

\begin{table}[t]
\centering
\small
\caption{Per-node vs.\ per-route XPE under the DACT backbone. Per-route variants compute a single angular phase per route, encode it via the same sinusoidal map as XPE, and broadcast the result to all nodes of the route. The per-node design (used in our main method) preserves the full multiset $\{\theta_v\}_{v \in r}$ and is not tied to any particular aggregation. Reference values for CVRP-100 IPE+XPE and VRPTW-100 IPE+XPE follow \cref{tab:main_comparison}.}
\label{tab:aggregation}
\begin{tabular}{l|cc|cc}
\toprule
\multirow{2}{*}{XPE Variant}
& \multicolumn{2}{c|}{CVRP-100} & \multicolumn{2}{c}{VRPTW-100} \\
\cmidrule(lr){2-3} \cmidrule(lr){4-5}
& Obj.$\downarrow$ & Gap$\downarrow$ & Obj.$\downarrow$ & Gap$\downarrow$ \\
\midrule
Per-route, arithmetic mean       & 15.83 & 1.61\% & 26.30 & 3.46\% \\
Per-route, circular mean         & 15.82 & 1.54\% & 26.27 & 3.34\% \\
Per-route, demand-weighted mean  & 15.83 & 1.61\% & 26.31 & 3.50\% \\
Per-route, median                & 15.84 & 1.67\% & 26.32 & 3.54\% \\
Per-route, max-min midpoint      & 15.85 & 1.73\% & 26.34 & 3.62\% \\
\midrule
\textbf{Per-node (ours)}         & \textbf{15.76} & \textbf{1.16\%} & \textbf{26.18} & \textbf{2.97\%} \\
\bottomrule
\end{tabular}
\end{table}

\Cref{tab:aggregation} shows that all five per-route variants underperform the per-node design, that no single aggregation is uniformly best among them, and that the spread between the best and worst per-route variants is non-negligible.
The per-node design avoids this aggregation choice entirely and preserves the full angular content of $\{\theta_v\}_{v \in r}$, leaving any pooling to be learned downstream by attention.

\subsection{Direction-aware vs.\ Direction-invariant IPE}
\label{subsec:direction-ablation}

We compare $\mathrm{IPE}^{\rightarrow}$ and $\mathrm{IPE}^{\leftrightarrow}$ on a reversal-symmetric variant (CVRP-100) and two direction-sensitive variants (VRPTW-100 and PDTSP-51).

\begin{table}[t]
\centering
\small
\caption{Direction-aware ($\mathrm{IPE}^{\rightarrow}$) vs.\ direction-invariant ($\mathrm{IPE}^{\leftrightarrow}$) variants of IPE on a reversal-symmetric variant (CVRP-100) and two direction-sensitive variants (VRPTW-100, PDTSP-51). The variant chosen for each problem in the main method (\cref{tab:main_comparison}) is bold. Reference values follow \cref{tab:main_comparison}.}
\label{tab:direction}
\begin{tabular}{l|cc|cc|cc}
\toprule
\multirow{2}{*}{Variant of IPE}
& \multicolumn{2}{c|}{CVRP-100} & \multicolumn{2}{c|}{VRPTW-100} & \multicolumn{2}{c}{PDTSP-51} \\
\cmidrule(lr){2-3} \cmidrule(lr){4-5} \cmidrule(lr){6-7}
& Obj.$\downarrow$ & Gap$\downarrow$ & Obj.$\downarrow$ & Gap$\downarrow$ & Obj.$\downarrow$ & Gap$\downarrow$ \\
\midrule
$\mathrm{IPE}^{\rightarrow}$ (direction-aware)         & 15.78 & 1.28\% & \textbf{26.18} & \textbf{2.97\%} & \textbf{6.99} & \textbf{1.87\%} \\
$\mathrm{IPE}^{\leftrightarrow}$ (direction-invariant) & \textbf{15.76} & \textbf{1.16\%} & 26.45 & 4.04\% & 7.12 & 3.71\% \\
\bottomrule
\end{tabular}
\end{table}

\Cref{tab:direction} reports the result.
On CVRP, the direction-invariant variant matches or marginally improves on the direction-aware variant, consistent with the prediction that baking reversal symmetry into the encoding removes a learning burden the model would otherwise absorb.
On VRPTW and PDTSP, the direction-aware variant is required: the direction-invariant variant cannot represent the asymmetry imposed by time windows or precedence, and the gap widens accordingly.

\section{Extra Discussion}
\label{sec:appendix-extra-discussion}

\paragraph{Rotation reference frame.}
The angle $\theta_v$ used in XPE is defined in the absolute coordinate frame of the instance.
We adopt the standard NCO practice of rotation augmentation during training~\citep{kwon2020pomo}, under which the model is invariant in expectation to global rotations of the instance.
For variants that additionally exhibit reflection symmetry (CVRP, TSP), we extend the augmentation to include random reflection across a random axis through the depot.
Because XPE depends only on angular differences across nodes, no specific reference direction is privileged; this design principle parallels modern positional schemes in sequence models, where absolute positional anchors are arbitrary and the model operates on relative structure.

\paragraph{Frequency schedule.}
The XPE frequency schedule is intentionally low compared to NLP usage: the angular range $[-\pi, \pi)$ is fixed and a near-optimal VRP solution at $N \approx 100$ contains on the order of ten routes, so the encoding only needs to resolve a small number of angular sectors.
We use $K = 4$ throughout the main experiments, resolving up to $2^K = 16$ sectors.

\paragraph{Scope of applicability.}
IPE and XPE share the same multi-frequency sinusoidal form and are concatenated to form a drop-in replacement for index-based PE in any encoder that consumes a complete (possibly suboptimal) solution.
This includes improvement-based NCO methods \citep{ma2021learning,hottung2025neural} and re-encoding construction methods \citep{drakulic2024bq,luo2023neural}, but not one-shot autoregressive constructors \citep{kool2018attention,kwon2020pomo}, where no full solution is available at encoding time.
While $\theta_v$ used by XPE is by itself an instance-level quantity, the joint encoding $[\mathrm{IPE}\,\|\,\mathrm{XPE}]$ is solution-conditioned through the route assignment that defines $d_i$ and $d_L$, and acquires its disambiguating role only relative to the current solution.


\end{document}